\pdfoutput=1

\documentclass[11pt]{article}
\usepackage{graphicx}
\usepackage{subcaption}
\usepackage{float}
\usepackage[preprint]{acl}
\usepackage{makecell}
\usepackage{multirow}
\usepackage{times}
\usepackage{latexsym}
\usepackage{enumitem}
\usepackage[T1]{fontenc}

\usepackage[utf8]{inputenc}

\usepackage{microtype}

\usepackage{inconsolata}

\usepackage{graphicx}
\usepackage{amsmath}
\usepackage{amssymb}
\usepackage{booktabs}
\usepackage{longtable}
\usepackage{enumitem}
\usepackage{tabularx}
\usepackage{graphicx}
\title{Deconstructing Instruction-Following: A New Benchmark for Granular Evaluation of Large Language Model Instruction Compliance Abilities}

\author{Alberto Purpura, Li Wang, Sahil Badyal, Eugenio Beaufrand, Adam Faulkner \\
  Card Intelligence, Capital One \\
  \texttt{\{alberto.purpura, li.wang, sahil.badyal,}\\
	\texttt{eugenio.beaufrand, adam.faulkner\}@capitalone.com}  \\}

\begin{document}
\maketitle
\begin{abstract}
Reliably ensuring Large Language Models (LLMs) follow complex instructions is a critical challenge, as existing benchmarks often fail to reflect real-world use or isolate compliance from task success. We introduce MOSAIC (MOdular Synthetic Assessment of Instruction Compliance), a modular framework that uses a dynamically generated dataset with up to 20 application-oriented generation constraints to enable a granular and independent analysis of this capability.
Our evaluation of five LLMs from different families based on this new benchmark demonstrates that compliance is not a monolithic capability but varies significantly with constraint type, quantity, and position. The analysis reveals model-specific weaknesses, uncovers synergistic and conflicting interactions between instructions, and identifies distinct positional biases such as primacy and recency effects. These granular insights are critical for diagnosing model failures and developing more reliable LLMs for systems that demand strict adherence to complex instructions.
\end{abstract}

\section{Introduction}

Large Language Models (LLMs) are increasingly employed as functional blocks within agentic pipelines and information processing systems \cite{anthropic2024introducing}. In this context, LLMs are expected to act as reliable components that receive data, perform an operation, and generate a response compliant with a set of standards. These standards are typically specified as instructions within the input prompt and do not necessarily overlap with the task the LLM has to complete. For example, an LLM might be tasked with summarizing a business document (the core task), while simultaneously being required to adhere to a list of instructions e.g., ``The summary must be under 100 words''. 
The model's success in such a system depends not only on the quality of its summary but also on its strict adherence to these additional constraints.
This paper proposes a novel perspective on the evaluation of LLM instruction compliance by introducing a new evaluation benchmark and metrics -- MOSAIC (MOdular Synthetic Assessment of Instruction Compliance).
Our approach is motivated by a closer examination of the current landscape, which reveals several gaps in existing evaluation and mitigation strategies.
First, evaluation benchmarks often utilize constraints that, while easily measurable, lack relevance to real-world application -- e.g., \textit{''In your response, words with all capital letters should appear at least/around/at most {N} times.''} from IFEval \cite{zhou2023instruction}. Second, many prominent benchmarks, such as InFoBench \cite{qin2024infobenchevaluatinginstructionfollowing}, evaluate instruction following with constraints that are specific to the current prompt task. This coupling makes it difficult to separately evaluate an LLM's intrinsic ability to follow \textit{instructions} from its ability to solve the \textit{task} itself. 
This highlights a critical distinction: task accuracy measures the factual correctness of the core output (e.g., providing the right answer to a question), whereas instruction following, or compliance, measures adherence to meta-rules about the output's format, style, or structure. A model can be accurate but non-compliant, or compliant but inaccurate. 
Third, evaluations are frequently conducted with a small number of constraints, largely overlooking how their inter-dependencies and quantity affect compliance. While ComplexBench \cite{wen2024benchmarkingcomplexinstructionfollowingmultiple} constitutes a notable exception by exploring constraint interactions, its analysis is still confined to 4-5 constraints at a time. 



MOSAIC is distinguished by its emphasis on \textit{modularity} and \textit{real-world applicability}. Instead of embedding instructions within a static prompt, we propose a dynamic framework where complex, application-oriented constraints are provided as a modular list.
We choose to evaluate our models on the constrained text generation task, considering 32 different variants. Unlike tasks with limited output spaces like classification, text generation allows for a more rigorous and fine-grained analysis of compliance and its open-ended nature provides a rich canvas for applying a diverse set of stylistic, structural, and semantic constraints. Furthermore, treating constraints as a modular list decouples them from the primary task. This modularity allows our evaluation framework to be generalized: the same set of constraints can be paired with numerous other tasks, enabling a pure assessment of the instruction-following capability, independent of the model's proficiency in a specific domain. 

Finally, our use of dynamic and modular dataset generation provides a critical defense against data leakage, a pervasive issue that compromises the validity of many public benchmarks.

\noindent Our contributions can be summarized as follows:
\begin{itemize}[nosep]
\item We introduce MOSAIC: a novel, modular benchmark for evaluating instruction following, featuring complex, application-oriented semantic constraints that are dynamically generated to robustly assess model capabilities.\footnote{\url{https://github.com/CapitalOne-Research/llm-instruction-following-compliance}.}
\item We propose a suite of evaluation metrics to measure instruction compliance, including methods for assessing pairwise constraint interactions, and constraint ordering effects.
\item We thoroughly evaluate different commercial and open-source LLMs on our benchmark, highlighting their differences relying on our evaluation framework.
\end{itemize}

\section{Related Work}
\label{sec:litrev}
Research into evaluating the instruction-following capabilities of LLMs has produced a variety of benchmarks and methodologies. These approaches can be broadly categorized by their strategies for \textit{instruction formation} (how complex instructions are constructed), their emphasis on \textit{instruction complexity} (the nature of the constraints applied), and their choice of \textit{evaluation metrics} (how compliance is measured).
One major line of work employs a ``bottom-up'' approach to instruction formation, where prompts are built by combining multiple simple and individually verifiable instructions. For instance, IFEval \citep{zhou2023instruction} assesses adherence to precise lexical and formatting rules, such as word counts or keyword mentions, that allow for exact, rule-based verification. IFEval-Extended \citep{kovalevskyi2024ifeval} advanced this method by introducing dynamic prompts with parameterized constraints to better defend against data leakage. While effective for measuring adherence to simple rules, these benchmarks often use constraints that are not representative of real-world applications and tend to examine constraints in isolation, overlooking their interactions. Our work builds on the benefits of this dynamic, bottom-up approach while incorporating more complex and interdependent constraints.

Another category of research focuses on more complex, application-grounded scenarios. Some works in this area adopt a 	``top-down'' method, where a high-level instruction is manually deconstructed into a set of verifiable sub-requirements. InfoBench \citep{qin2024infobenchevaluatinginstructionfollowing} is a key example, using LLM-as-a-Judge to score satisfaction on these sub-requirements through Yes/No questions. Similarly, CELLO \citep{he2024can} evaluates compliance with constraints grounded in real-world tasks using rule-based scoring. While these benchmarks increase the practical relevance of the instructions, they typically do not provide a systematic analysis of how composing multiple constraints affects model performance and the evaluation on the compliance of an LLM is dependent on the task in each prompt instead of a fixed list of constraints.

A third group of studies directly investigates the impact of instruction composition. FollowBench \citep{jiang2024followbenchmultilevelfinegrainedconstraints} takes an incremental, bottom-up approach by progressively adding constraints to observe their effect on compliance, but it does not formalize the nature of constraint interactions. ComplexBench \citep{wen2024benchmarkingcomplexinstructionfollowingmultiple} provides a more direct analysis by explicitly composing constraints with logical operators and evaluating performance under these composite scenarios. Our research is heavily inspired by this focus on constraint interaction, and we extend it by methodically exploring the effects of constraint ordering and conflicts, 
which we find to be critical factors. Finally, a comprehensive survey by Garbacea et al. \citep{garbacea2022constrained} confirms that constrained text generation remains a significant challenge, highlighting the difficulties models face in controlling outputs to meet specific conditions.

\begin{table*}[h]
\resizebox{\linewidth}{!}{%
    \centering
    \scriptsize 
    \renewcommand{\arraystretch}{1.3} 
    \begin{tabularx}{\textwidth}{@{} l >{\raggedright\arraybackslash}p{2.5cm} c >{\centering\arraybackslash}p{2cm} >{\centering\arraybackslash}p{2cm} c >{\raggedright\arraybackslash}p{2.5cm} @{}}
        \toprule
        \textbf{Benchmark} & \textbf{Task type} & \textbf{\makecell{No. of \\ prompts}} & \textbf{\makecell{Avg. prompt \\ length}} & \textbf{\makecell{No. of \\ constraint types}} & \textbf{\makecell{Avg. no. of constraints \\ per prompt}} & \textbf{\makecell{Compliance \\ analysis}} \\
        \midrule
        IFEval & General writing tasks including poems, songs, essays, blog posts, etc. & 541 & 37.07 words & 9 groups and 25 types & 1.54 & Prompt- and constraint-level compliance \\
        
        InfoBench & Tasks from 143 domains including engineering, arts, economics, etc. & 500 & 38.35 words & 5 types & 4.50 & Decomposed constraint-level compliance \\
        
        CELLO & 9 complex NLP tasks including extraction, planning, meta-prompt, etc. & 523 & N/A & 4 types & N/A & Prompt-level and fine-grained constraint-level compliance \\
        
        FollowBench & 50+ NLP tasks including closed- and open-ended questions & 820 & 274.77 words & 5 types & 3.00 & Prompt-level and (consecutive) constraint-level compliance \\
        
        ComplexBench & Tasks derived from real-world application scenarios and open-source instruction following benchmarks (IFEval, Infobench, Followbench) & 1150 & 477.51 chars & 4 groups and 19 types & 4.19 & Composition-based prompt-level and question-based constraint-level compliance \\
        \midrule
        
        \textbf{MOSAIC} & \textbf{4 types of text-based content generation tasks with 8 types of products/service} & \textbf{4000} & \textbf{222.10 words, 1608.72 chars} & \textbf{5 groups and 21 types} & \textbf{10.50} & \textbf{Prompt-level compliance. Single, pairwise and position-based constraint compliance} \\
        \bottomrule
    \end{tabularx}}
    \caption{Comparison of the proposed approach to other benchmarks in the literature.}
    \label{tab:benchmarks}
\end{table*}

Table \ref{tab:benchmarks} contrasts MOSAIC with the related works discussed above \citep{zhou2023instruction, qin2024infobenchevaluatinginstructionfollowing, he2024can, jiang2024followbenchmultilevelfinegrainedconstraints, wen2024benchmarkingcomplexinstructionfollowingmultiple} across six key dimensions, including task diversity, scale, and constraint density. MOSAIC significantly outperforms existing datasets in two critical areas: it offers the largest scale (4,000 prompts) and the highest complexity, averaging 10.5 constraints per prompt -- more than double the next highest competitor. Additionally, our modular synthetic generation pipeline ensures the dataset is highly scalable, allowing for easy adjustments to both size and complexity to meet future research needs.
\section{Proposed Approach}

\subsection{Dataset Generation}
To evaluate the instruction following ability of LLMs, we dynamically assemble a synthetic dataset where prompts are generated as a concatenation of (i) a text generation task (Table \ref{tab:content}), (ii) product or service (Table \ref{tab:products}), and (iii) one or more constraints (Table \ref{tab:constraints}), in the order. To combine them, we use the following prompt template.\footnote{The shown prompt template is using Llama-style special tokens, we update the special tokens to delimit system/user/assistant messages according to the defaults of each model used for experimentation.}

\noindent \texttt{\small 
<|begin\_of\_text|>\\
<|start\_header\_id|>system<|end\_header\_id|>\\
You are a writing assistant. Your task is to <taskDescription>.\\
Ensure your draft complies with all of the following requirements:\\ <constraintsList>\\
<|eot\_id|>\\
<|start\_header\_id|>user<|end\_header\_id|>\\
Product/Service: <prodServDescription>.<|eot\_id|>\\
<|start\_header\_id|>assistant<|end\_header\_id|>
}

\begin{figure*}
\centering
\includegraphics[width=\linewidth
]{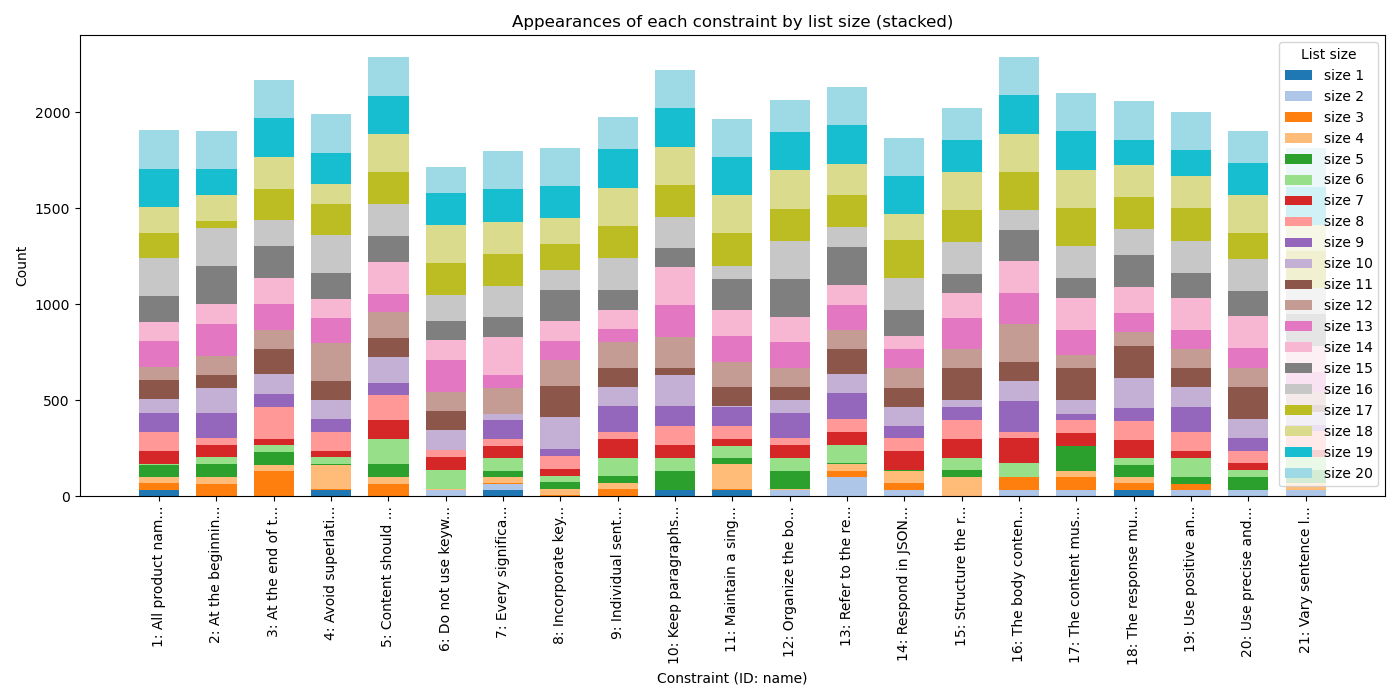}
\caption{Distribution of constraint appearances by list size. Each bar shows the total number of times each constraint appears in prompts with a given number of constraints, illustrating how constraint usage varies across different list sizes in the dataset.}
\label{fig:prompts_distrib}
\vspace{-1em}
\end{figure*}

\begin{table}[h!]
\centering
\small
\begin{tabularx}{0.5\textwidth}{X}
\toprule
\textbf{Content Type and Description} \\
\midrule
\textbf{Marketing email:} Write a marketing email to promote a given product or service. \\
\midrule
\textbf{Product review:} Write a detailed description and review of a given product or service. \\ \midrule

\textbf{FAQ Entry:} Write a clear and concise entry for a product's Frequently Asked Questions page, explaining a specific feature.
\\ \midrule

\textbf{Internal Memo:} Write a brief internal memo for employees announcing a new product or service. \\ 
\bottomrule
\end{tabularx}
\caption{Content generation tasks.}
\label{tab:content}
\vspace{-1em}
\end{table}

\begin{table}[h!]
\centering
\small
\begin{tabularx}{0.5\textwidth}{X} 
\toprule
\textbf{Product/Service Type and Description} \\
\midrule
\textbf{Smartphone:} A flagship smartphone featuring a 120Hz dynamic display, a triple-lens camera system with a 108MP main sensor, and 2-day battery life. \\ \midrule
\textbf{Wireless Earbuds:} True wireless stereo earbuds with active noise cancellation, 8-hour playback on a single charge, and a water-resistance rating of IPX7. \\
\midrule
\textbf{Savings Account:} An online savings account with a 4.5\% Annual Percentage Yield (APY), no monthly maintenance fees, and FDIC insurance up to \$250,000. \\ \midrule
\textbf{Credit Card:} A credit card offering 3\% cashback on rotating categories (gas, groceries), 1\% on all other purchases, and no annual fee for the first year. \\
\midrule
\textbf{Meditation App:} A subscription-based mobile app offering a library of over 500 guided meditations, sleep stories, and mindfulness exercises. \\ \midrule
\textbf{Fitness Monitor:} A wrist-worn device that tracks steps, heart rate, sleep patterns, and SpO2 levels, with a companion app for goal setting and progress monitoring. \\
\midrule
\textbf{Meals Subscription:} A weekly subscription box that delivers pre-portioned ingredients and recipes for chef-designed meals, with options for various dietary needs. \\ \midrule
\textbf{News Streaming:} An online service providing 24/7 live access to international news channels, documentaries, and in-depth political analysis. \\

\bottomrule
\end{tabularx}
\caption{Product or Service descriptions.}
\label{tab:products}
\vspace{-1em}
\end{table}

\begin{table*}[h!]
\centering
\small 
\begin{tabularx}{\textwidth}{X} 
\toprule
\textbf{Constraint Type and Description} \\
\midrule

\makecell[X]{ \textbf{Formatting:} \\ 
-- \textbf{Keep paragraphs short:} Keep paragraphs short, ideally 2-4 sentences.\\ 
-- \textbf{Keep body short:} The body content should be organized into 2-3 paragraphs.\\
-- \textbf{Structure body into lists:} Organize the body content into lists using only dashes.\\
-- \textbf{Include <BOC> token:} At the beginning of the generated content, include the special token <BOC>.\\
-- \textbf{Include <EOC> token:} At the end of the generated content, include the special token <EOC>.\\
-- \textbf{Respond in JSON: } Respond in JSON format following the schema: \texttt{\{"response": <your response>\}}. 
} \\

\midrule

\makecell[X]{\textbf{Lexical:} \\
-- \textbf{Flesch Reading Ease 70-80: } Content should target a Flesch Reading Ease 
\cite{flesch1948new} level between 70 and 80 to ensure broad accessibility. \\ 
-- \textbf{Use positive language:} Use positive and empowering language (e.g., 'opportunity', 'benefit', 'simplify') and avoid negative or fear-based terms (e.g., 'problem', 'risk', 'failure'). \\
-- \textbf{Use given keywords: } Incorporate keywords aligned with the brand voice i.e., <kw1, kw2, ..>. \\
-- \textbf{Avoid given keywords: }  Do not use keywords like <{kw1}, {kw2}, ..>. \\
-- \textbf{Use custom variable :} Refer to the recipient of the message using the variable \{\{FirstName\}\} enclosed by double curly brackets. \\
}\\
\midrule
\makecell[X]{\textbf{Syntactic:} \\
-- \textbf{Vary sentence length: } Vary sentence length and structure to create a compelling rhythm. Mix simple, compound, and complex sentences. \\
-- \textbf{Keep sentences short:} Individual sentences should not exceed 25 words to maintain clarity and momentum. \\
}\\
\midrule

\makecell[X]{\textbf{Semantic:} \\
-- \textbf{Use given tone: } Maintain a single, consistent tone throughout the entire response as specified. Your response tone should be: <SelectedTone>.\\
-- \textbf{Use inverted pyramid principle: } Structure the response following the "inverted pyramid" principle. The most critical piece of information (the core answer, main announcement, or key takeaway) must be presented at the beginning, before supporting details or secondary context. \\
-- \textbf{Avoid contradictions: } The response must not contain any internal contradictions. All stated facts, arguments, and data points must be consistent with each other from start to finish. \\
--  \textbf{Substantiate every claim: } Every significant claim, benefit, or conclusion must be supported by a reason or piece of evidence within the text. Avoid making unsupported assertions. For example, instead of "It's faster," write "It's faster because it uses a next-generation processor." \\
-- \textbf{Clear purpose: }The content must directly address the primary underlying question of the target audience for the task (e.g., 'How will this help me?', 'What do I need to know?', 'Is my problem solved?'). The purpose of the communication must be clear.
}\\\midrule

\makecell[X]{\textbf{Business/Legal:} \\
-- \textbf{Use unambiguous language: } Use precise and unambiguous language. Avoid vague terms or phrases that could be misinterpreted by the target audience. All instructions, descriptions, or conclusions should be explicit and clear. \\
--  \textbf{Report correct features: } All product names, features, and numerical data (e.g., prices, percentages) must be accurate and up-to-date as of the generation date. \\
-- \textbf{Avoid unsubstantiated superlatives: } Avoid superlatives (e.g., 'best', 'greatest') unless they can be substantiated by a verifiable source, which must be cited or linked. \\
} \\
\bottomrule
\end{tabularx}
\caption{Generation constraints, categorized by formatting, lexical, syntactic, semantic, and business/legal categories. \\ We report the <SelectedTone> and <kw $n$> values we considered in Appendix \ref{sec:appendix}.}
\label{tab:constraints}
\vspace{-1em}
\end{table*}

We design the dataset generation process in two phases: first, the creation of a comprehensive, large-scale collection of prompts, and second, the extraction of a smaller, statistically balanced subset for controlled analysis.

\textbf{Comprehensive Generation.}
We begin by generating an initial comprehensive dataset by creating prompts containing every possible pairing of a content generation task (Table \ref{tab:content}) and a product/service (Table \ref{tab:products}). For each of these pairings, we introduce a variable number of constraints, from 1 to 20, drawn (without replacement) from a pool of 21 unique constraints (Table \ref{tab:constraints}).
To mitigate potential biases related to the order in which instructions are presented, we also control for constraint permutation. For constraint sets of size 3 or fewer, we generate all possible permutations for each combination. For larger constraint sets, where generating all permutations would be computationally infeasible, we generate multiple random shuffles for each combination. Finally, we dynamically populate certain constraints containing placeholders, such as <SelectedTone> or keywords, with values appropriate for the given product context during prompt assembly. This initial process results in a very large dataset containing a highly diverse range of 765,472 prompts.

\textbf{Stratified Sampling.}
The initial generation process yields a dataset that is too large for efficient evaluation and inherently unbalanced in its distribution of constraints list sizes. To address this problem, we create a smaller, balanced subset through a stratified sampling procedure.
Our primary objective is to achieve a uniform distribution of each of the 21 constraints across all list sizes (1 to 20), across all possible ranks (i.e., positions) within those lists and across each task and product/service. This allows for a clean analysis of how constraint count and position affect model performance, independent of the specific constraint content, task or product/service variant.
To construct the final subset, we sample a total of 4000 prompts from each pool of equivalent prompts grouped by task description, product/service description and constraints list size (1 to 20). Note that we cannot directly optimize for the rank at which each constraint appears in its respective list in a prompt since we perform sampling at the prompt level. Nonetheless, this stratified approach yields a dataset that remains relatively balanced also in terms of the ranks at which each constraint appears, as visualized in Figure \ref{fig:prompts_distrib}. 
This balance is methodologically critical, as it decouples the influence of a constraint's content from its position and other factors, thereby enabling an unbiased analysis of instruction order.

\subsection{Evaluation Metrics}

We employ several evaluation measures to evaluate the instruction following abilities of a model.

\noindent \textbf{Single and Pairwise Constraint Compliance.} These measures allow us to gauge the difficulty of single constraints independently from the others and to measure the interaction strength of different constraints from the perspective of an LLM. The Single Constraint Compliance (SCC) measure is computed as:
\begin{equation}
\texttt{SCC}_c = \frac{\sum_{r \in R}\texttt{is\_followed}(c, r)}{|R|}, \forall c \in C,
\end{equation} 

where $\texttt{SCC}_c$ is the rate at which constraint $c \in C$ is followed in all of the considered responses $r \in R$. Differently from $SCC$, we propose a novel Pairwise Constraint Compliance ($PCC$) metric computed as:
\begin{multline*}
\texttt{PCC}_{c_i, c_j} = \frac{\sum_{r \in R}\texttt{is\_followed}(c_{i}, c_{j}, r)}{|R|},\\ \forall c_i, c_j \in C,
\end{multline*} 
where $c_i, c_j \in C$ are two distinct constraints that should be followed by an LLM when providing a given response $r\in R$.

\noindent \textbf{Position-based Constraint Compliance.} The compliance rate associated with each constraint may be affected by its position in the prompt. We propose this novel evaluation measure to assess whether there are any correlations between the position of a constraint in a list, its size and the SCC measure associated to that constraint. We compute this measure as follows:
\begin{equation}
\texttt{PosCC}_p = \frac{\sum_{r \in R_p}\texttt{is\_followed}(c_{r, p}, r)}{|R_p|},
\end{equation}
where p is the position in the list of constraints (e.g., 1st, 2nd, 3rd), $R_p$ is the set of all responses to prompts containing at least p constraints (i.e., $R_p =\{r \in R: | C_r | \geq p\}$), and $c$, $r$, $p$ indicate the specific constraint $c$ at position $p$ for a given response $r$. This metric allows us to compute the constraint compliance as a function of position to reveal any potential biases, such as models paying less attention to constraints listed earlier or later in a prompt.

\noindent \textbf{Prompt-level instruction following accuracy.} Since an LLM request may come with multiple constraints, this metric -- originally proposed in \cite{kovalevskyi2024ifeval} -- provides an estimate for the number of perfectly compliant responses that we can expect to receive from a given LLM. Given a response $r$ and its respective set of constraints $C_r$, the respective Prompt following Accuracy ($PA_r$) can be calculated with the following formula:

\begin{equation}
\texttt{PA}_r = \frac{\sum_{c \in C_r}\texttt{is\_followed}(c, r)}{|C|},
\end{equation} 
where for each response $r$ to a prompt, we compute the percentage of constraints that are respected. In our evaluation, we also report the average of these values across all LLM responses for a broader comparison.


%
\subsection{Evaluation Strategy}
\label{sec:eval_strategy}
To evaluate whether a response complies with each of the constraints indicated in its prompt we adopt the following approach. For the Formatting, Lexical and Syntactic constraints reported in Table \ref{tab:constraints}, we developed different evaluation functions relying on the NLTK library.\footnote{\url{https://www.nltk.org/}}
Each of these functions returns a value between 0.0 and 1.0, indicating the rate of compliance to each of the constraints -- 0.0 indicates a complete failure to comply with the constraint while 1.0 indicates a perfect compliance. More details on these functions' implementation are available in Appendix \ref{sec:appendix_eval_func}. For what concerns the more complex Semantic and Business/Legal constraints, we rely on an LLM-as-a-judge approach. We prompt a GPT-4o-mini model with a different prompt each time to judge whether a response complies with a given constraint. The model is tasked with returning a score from 1 to 10 to judge how much the response complies with each applicable constraint reported in its prompt (one at a time).
To compute our final evaluation metrics, all of the aforementioned fractional scores are converted to a binary value (0 or 1) if they are $<0.5$ or $\geq0.5$, respectively. More details on the prompts used to verify each constraint are provided in Appendix \ref{sec:appendix_eval_func}.

To validate the reliability of our LLM-as-a-judge for constraint verification, we conduct a human evaluation study. In this study, two expert human annotators evaluated the compliance of a randomly selected set of 250 model responses. This sample was constructed to cover the full range of constraints that the LLM-as-a-judge assesses.
We asked each expert to provide a binary compliance score for a specific constraint, given the original prompt and the model-generated response. We then calculated the agreement scores. The agreement between the LLM-as-a-judge and the human annotators was 0.82 (vs. Annotator 1) and 0.83 (vs. Annotator 2), with an inter-human agreement of 0.94. These figures demonstrate a high degree of reliability, confirming the LLM-as-a-judge's efficacy for our scenario.

\section{Evaluation Results}

\begin{table*}[h]
\small
\centering
\resizebox{\textwidth}{!}{%
\begin{tabular}{p{4.5cm} r r r r r r r}
\toprule
\textbf{Constraint} &  \textbf{Llama-3.1-8B} & \textbf{Llama-3.3-70B} & \textbf{Qwen3 8B} & \textbf{Mixtral-8x-7B} & \textbf{Deepseek-R1 8B} & \textbf{Gemini 2.5 Flash} & \textbf{Claude 3.7 Sonnet}\\
\midrule
Avoid contradictions & 1.0000 & 1.0000 & 1.0000 & 1.0000 & 1.0000 & 0.9995 & 0.9917 \\
Avoid given keywords & 0.8218 & 0.8085 & 0.9059 & 0.7481 & 0.9112 & 0.9994 & 0.9959 \\
Avoid unsubstantiated superlatives & 0.7289 & 0.6898 & 0.7860 & 0.7295 & 0.6601 & 0.7052 & 0.7461 \\
Clear purpose & 1.0000 & 1.0000 & 0.9995 & 0.9900 & 0.9990 & 0.9995 & 0.9967 \\
Flesch reading ease 70-80 & 0.7735 & 0.8053 & 0.6798 & 0.7765 & 0.4174 & 0.5395 & 0.6024 \\
Include <BOC> token & 0.9421 & 0.9963 & 1.0000 & 0.8997 & 0.9221 & 1.0000 & 1.0000 \\
Include <ECO> token & 0.9542 & 0.9756 & 0.9550 & 0.8999 & 0.8489 & 0.9981 & 0.9995 \\
Keep body short & 0.7608 & 0.8080 & 0.9933 & 0.5715 & 0.9450 & 0.9917 & 0.8483 \\
Keep paragraphs short & 0.8721 & 0.8806 & 0.6206 & 0.9059 & 0.8736 & 0.6401 & 0.8158 \\
Keep sentences short & 0.9990 & 0.9995 & 0.9976 & 0.9985 & 0.9958 & 0.9975 & 0.9995 \\
Report Correct Features & 0.9995 & 0.9974 & 0.9986 & 0.9261 & 0.9754 & 0.9963 & 0.9916 \\
Respond in JSON & 0.9978 & 1.0000 & 0.9961 & 0.9914 & 0.9723 & 0.9995 & 0.9989 \\
Structure body into lists & 0.9689 & 0.9864 & 0.3359 & 0.7473 & 0.0699 & 0.4220 & 0.6631 \\
Substantiate every claim & 0.7902 & 0.7964 & 0.9277 & 0.8103 & 0.8022 & 0.7765 & 0.9333 \\
Use Custom Variable & 0.8532 & 0.8386 & 0.9396 & 0.8907 & 0.8691 & 1.0000 & 0.9939 \\
Use given keywords & 0.9037 & 0.8935 & 0.9761 & 0.9367 & 0.8825 & 0.9917 & 0.9444 \\
Use given tone & 0.9990 & 0.9995 & 0.9979 & 0.9944 & 0.9974 & 0.9990 & 0.9924 \\
Use inverted pyramid principle & 0.9975 & 0.9980 & 0.9987 & 0.9872 & 0.9765 & 0.9816 & 0.9936 \\
Use positive language & 1.0000 & 0.9995 & 1.0000 & 0.9970 & 0.9995 & 1.0000 & 0.9955 \\
Use unambiguous language & 1.0000 & 1.0000 & 1.0000 & 0.9911 & 0.9986 & 0.9995 & 0.9937 \\
Vary sentence length & 0.9945 & 0.9972 & 0.9605 & 0.9619 & 0.9627 & 0.8719 & 0.9890 \\
\bottomrule
\end{tabular}}
\caption{Single Constraint Compliance (SCC) rates for different constraints and models.}
\label{tab:scc_scores}
\vspace{-1em}
\end{table*}

We evaluate the ability of LLMs of different sizes to comply with the list of constraints indicated in Table \ref{tab:constraints} i.e., Llama 3.3 70B, Llama 3.1 8B \cite{grattafiori2024llama}, Deepseek-R1 8B \cite{liu2024deepseek}, Qwen3 8B \cite{bai2023qwen}, Claude 3.7 Sonnet, Gemini Flash 2.5 and 
Mixtral-8x-7b \cite{jiang2024mixtral}. For Deepseek-R1 8B and Qwen 3 8B, we use Ollama\footnote{\url{https://ollama.com/}} to host an inference server with Q4\_K\_M quantization and its default parameters. For Llama 3.3 70B, Llama 3.1 8B and Mixtral-8x-7b, we use Huggingface Text Generation Inference (TGI)\footnote{\url{https://huggingface.co/docs/text-generation-inference/en/index}} for inference and generate their responses in a deterministic setting with Temperature set to 1.0. For Claude\footnote{\url{https://platform.claude.com/docs/en/about-claude/models/overview}} and Gemini\footnote{\url{https://docs.cloud.google.com/vertex-ai/generative-ai/docs/models/gemini/2-5-flash}} models we use the proprietary APIs with default settings.\footnote{Experiments were performed in November 2025.}
To avoid biases towards any of the models' responses, we use \texttt{gpt-4o-mini}\footnote{\url{https://openai.com/index/gpt-4o-mini-advancing-cost-efficient-intelligence/}} as our LLM-as-a-judge model.

\subsection{Single Constraints Compliance Analysis}
In Table~\ref{tab:scc_scores}, we report the $SCC$ scores for each of the 21 constraints across the evaluated models. The results reveal significant variance in compliance, not only between different models but also across different types of constraints, underscoring that instruction-following is not a monolithic capability. We observe that a subset of constraints, primarily related to high-level semantic coherence and style, are followed with near-perfect accuracy by all models. For instance, constraints such as \textit{Avoid contradictions}, \textit{Clear purpose}, \textit{Use positive language}, and \textit{Use unambiguous language} consistently achieve SCC scores approaching 1.0. This suggests that current models are well-aligned with fundamental principles of clear and consistent communication. Similarly, basic syntactic rules like \textit{Keep sentences short} are handled with high fidelity. Conversely, several constraints prove to be universally challenging. Adherence to a specific \textit{Flesch reading ease 70-80} score is notably difficult, with scores ranging from 0.80 down to a low of 0.42 for Deepseek-R1, indicating a difficulty in finely controlling text complexity. Likewise, the constraint to \textit{Avoid unsubstantiated superlatives} presents a consistent challenge, likely because it requires a form of self-critique and reasoning about evidence that is not yet robustly developed. Perhaps most strikingly, the results highlight critical model-specific weaknesses. The instruction to \textit{Structure body into lists} using dashes reveals a dramatic performance gap: the Llama 3 models achieve near-perfect compliance (0.97-0.99), whereas Qwen3 struggles (0.34) and Deepseek-R1 almost completely fails (0.07). This points to significant differences in how models interpret or prioritize specific structural formatting rules. The Llama 3 family of models demonstrates the most consistent and robust performance across the board, while other models like Qwen3 and Deepseek-R1 exhibit more specialized or ``spiky'' capability profiles, excelling on some constraints while failing significantly on others. These findings emphasize the necessity of granular, constraint-level evaluation, as high-level benchmarks can obscure critical, task-specific model failures. It is worth noting that the performance of commercial models such as Gemini 2.5 Flash or Claude 3.7 Sonnet is aligned to smaller open source ones.

In Table \ref{tab:scc_cat_scores} (in Appendix \ref{appendix:ccm}), we report the aggregated compliance rates by category. We observe that models consistently excel at Syntactic and Semantic constraints, which aligns with our finding that high-level coherence rules are well-followed. However, the category-level scores also highlight specific areas of weakness. For instance, the lower Formatting scores for Qwen3 (0.8168) and Deepseek-R1 (0.7720) directly reflect their significant struggles with the \textit{Structure body into lists} instruction, as noted earlier. This aggregated view confirms the overall performance trends, while reinforcing the need for the granular analysis in Table \ref{tab:scc_scores} to diagnose precise failure points.

\subsection{Pairwise Constraint Compliance Analysis}
To investigate the inter-dependencies between instructional constraints, we compute the Pearson correlation coefficient on the $PCC$ measurements obtained for each model. This analysis reveals statistically significant relationships, highlighting pairs of instructions that are either synergistic (positively correlated) or conflicting (negatively correlated) from the perspective of the language model. The most significant correlations highlighted by our metrics -- with an absolute value of the Pearson correlation coefficient ($r$) greater than 0.2 and p-value $< 0.05$ -- are reported in Tables \ref{tab:llama8b_pwise},  \ref{tab:llama70b_pwise},  \ref{tab:qwen_pwise}, \ref{tab:claude_pwise}, \ref{tab:gemini_pwise}, \ref{tab:mixtral_pwise} and \ref{tab:deepseek_pwise} in Appendix \ref{appendix:pwise}. 

\textbf{Consistent Synergies (Positive Correlations).}
Across the evaluated models, certain constraints demonstrate a strong synergistic relationship. A primary example is the correlation between instructions for clarity and precision. As detailed in Tables \ref{tab:mixtral_pwise} and \ref{tab:deepseek_pwise}, \textit{Clear purpose} and \textit{Use unambiguous language} are strongly and positively correlated for Mixtral ($r=0.575$) and especially for Deepseek-R1 ($r=0.866$). This suggests that for these models, achieving a clear purpose is intrinsically linked to the use of precise language. Similarly, a synergy exists between factuality and evidence. The constraints \textit{Report Correct Features} and \textit{Substantiate every claim} show a consistent positive correlation (e.g., $r=0.404$ for Mixtral-8x-7B, Table \ref{tab:mixtral_pwise}), indicating that prompts for factuality also encourage the model to provide supporting evidence.

\textbf{Consistent Conflicts (Negative Correlations).}
The analysis also identifies clear trade-offs where adherence to one constraint compromises adherence to another. The most prominent conflict exists between readability and lexical constraints. For nearly all models, there is a significant negative correlation between achieving a target \textit{Flesch reading ease 70-80} and being required to \textit{Use given keywords}. For instance, Llama 3.3 70B ($r=-0.278$, \ref{tab:llama70b_pwise}) and Mixtral ($r=-0.320$, Table \ref{tab:mixtral_pwise}) both struggle with this pair. This highlights a fundamental tension: forcing the model to use specific, and often more complex, keywords directly interferes with its ability to generate simple, readable text.

\textbf{Model-Specific Behaviors.}
Beyond general patterns, the models exhibit unique correlational profiles that likely reflect their distinct training. For example, Qwen3 demonstrates a strong positive correlation between \textit{Keep paragraphs short} and \textit{Keep body short} ($r=0.783$), but a strong negative correlation between \textit{Include <BOC> token} and \textit{Respond in JSON} ($r=-0.531$) (Table \ref{tab:qwen_pwise}). This suggests a structural conflict where generating a JSON object disrupts the model's adherence to adding a beginning-of-content token. By contrast, Mixtral-8x-7B shows a web of positive correlations centered on \textit{Use given tone}, linking it to numerous other qualitative constraints like \textit{Clear purpose} and \textit{Avoid contradictions} (Table \ref{tab:mixtral_pwise}). This may indicate that for Mixtral, establishing a specific tone favors more high-quality, compliant writing.

\subsection{Compliance by Constraint Position and Quantity}

\begin{figure}[h]
\includegraphics[width=\linewidth]{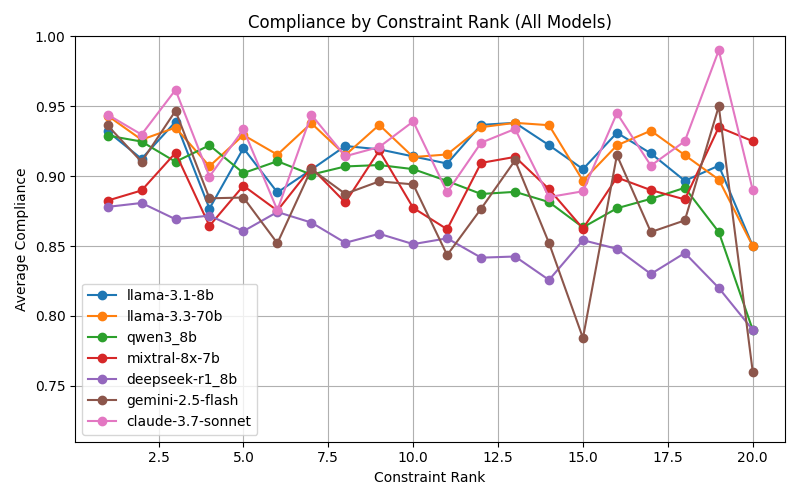}
\caption{Constraint compliance by rank.}
\label{fig:rank_compliance}
\vspace{-1em}
\end{figure}

\begin{figure}[h ]
\includegraphics[width=\linewidth]{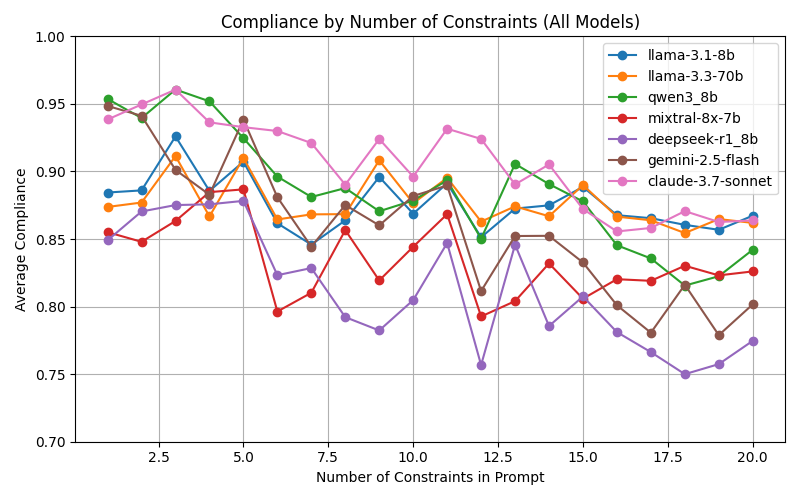}
\caption{Prompt compliance by number of constraints.}
\label{fig:prompt_compliance}
\vspace{-1.5em}
\end{figure}

To understand how models handle multiple constraints in a prompt, we analyze compliance based on their position and total number. We observe that the position of a constraint within a list significantly impacts its likelihood of being followed. As shown in Figure \ref{fig:rank_compliance}, most models exhibit a primacy effect: constraints placed at the beginning of a list are adhered to more reliably than those at the end. This trend is particularly evident for the Llama, Qwen3, and Deepseek models. By contrast, Mixtral-8x-7b, Gemini and Claude models demonstrate a recency effect, showing higher compliance with constraints that appear later in the prompt. More specifically, Claude and Gemini displayed a unique attention pattern compared to other models, with a very high compliance spike for constraints at index 19 and an abrupt compliance decrease when the constraint position increased to rank 20. The trends we observe are aligned with \citet{liu2023lost}'s paper, where the authors also observed changes in a model's performance when varying the position of key information in an LLM's input. 
We hypothesize that, in these cases, the differing positional biases stem from the models' core architectural differences. The recency effect observed in Mixtral-8x-7B Claude and Gemini may indirectly related to some architecture similarities -- e.g. the reliance on the mixture of experts architecture -- that we unfortunately cannot verify due to the close source of some of the models. 
We also examine how the total number of constraints affects a model's overall compliance rate. Figure \ref{fig:prompt_compliance} reveals distinct performance patterns based on prompt complexity. For prompts with few constraints (approximately 1-6), Qwen3 emerges as the most dependable model, consistently achieving the highest average compliance. As the number of constraints increases into the range of 7 to 15, performance becomes less predictable across all models. The significant fluctuations in this zone suggest that compliance becomes highly dependent on the specific nature and interaction of the constraints, rather than just their quantity. When prompts contain more than 15 constraints, nearly all models show a distinct drop in compliance. It is noteworthy that this performance degradation is more pronounced for models designed for reasoning, such as Deepseek-R1 8b and Qwen3, as the task complexity rises.

\section{Conclusions}
We introduced MOSAIC (MOdular Synthetic Assessment of Instruction Compliance), a novel modular benchmark for evaluating LLM instruction compliance that addresses critical limitations in existing literature. Unlike prior benchmarks that often conflate task success with compliance or use artificial constraints, our approach decouples these aspects by using a modular list of complex, application-oriented constraints. This allows for a more pure and realistic assessment of a model's intrinsic instruction-following capabilities. Furthermore, our research goes beyond prior work on constraint composition by providing a more granular analysis of how compliance is affected by constraint quantity, interaction, and crucially, ordering. Our analysis uncovered novel, model-specific positional biases, such as primacy and recency effects, as well as synergistic and conflicting relationships between instructions. By deconstructing compliance into these distinct factors, our work provides a powerful and generalizable diagnostic methodology for identifying specific failure points, offering critical insights for the development of more reliable and controllable LLMs and for a more targeted prompt engineering process.

\section*{Limitations}

While our study introduces a robust framework for evaluating instruction compliance, it is important to acknowledge its limitations.

First, our evaluation of complex semantic and business/legal constraints relies on an LLM-as-a-judge approach. The reliability of these evaluations is therefore dependent on the capabilities and potential biases of the judge model itself. While we employed a highly capable model and benchmarked it against human annotators, this method is not infallible and may not capture all nuances of compliance perfectly. Additionally, different constraints may not be always objectively gradable based on the use case.

Second, the benchmark, while comprehensive and dynamically generated to prevent data leakage, is synthetic. The constraints and tasks were designed to be representative of real-world applications, but they may not encompass the full spectrum of complexity, ambiguity, and implicit context found in human-generated instructions. The generalizability of our findings to entirely different domains (e.g., creative writing, scientific analysis, or code generation) remains to be explored.

Third, our analysis is based on a specific set of language models and a limited number of content generation tasks. The observed compliance patterns and the effectiveness of our proposed prompt refinement strategies may vary with different model architectures, sizes, or fine-tuning methods. Further research is needed to assess how these findings apply to other models, particularly larger, proprietary systems.
Finally, we would like to point out that even though our evaluation dataset contains prompts in English, our benchmark is uniquely adaptable to be automatically translated in different languages to allow for the instruction following abilities evaluation in other languages.



\bibliography{custom}

\appendix

\section{Appendix: Additional Related Works Considerations}

While the benchmarks in Table \ref{tab:benchmarks} are directly comparable to our work in their focus on single-turn, prompt-level compliance, other streams of research investigate instruction following along dimensions that are orthogonal to our study. These works address specific challenges such as conversation depth, context scaling, and source hierarchy. While they do not directly evaluate the granular interactions of dense constraints within a single prompt -- the primary contribution of our framework -- they highlight complementary challenges in the broader landscape of model compliance.

Unlike our focus on dense, single-turn interactions, {StructFlowBench} \citep{li-etal-2025-structflowbench} and {Multi-IF} \citep{he2024multiifbenchmarkingllmsmultiturn} examine compliance over conversation history, testing structural dependencies and memory degradation. Similarly, {LIFBench} \citep{wu-etal-2025-lifbench} identifies attention failures in massive context windows (up to 128k tokens). These works isolate the difficulty of maintaining state over time or distance, whereas our work isolates the difficulty of satisfying complex, simultaneous logic in the immediate turn.

Research also targets the dimensions of source authority and language. {IHEval} \citep{zhang-etal-2025-iheval} addresses the instruction hierarchy, focusing on adversarial conflicts between system and user prompts (jailbreak resistance) rather than the logical constraint conflicts we analyze. Finally, {M-IFEval} \citep{liu-etal-2025-maxife} demonstrates that compliance capabilities vary significantly across languages, highlighting that current English-centric benchmarks may overestimate general model robustness. These studies underscore that granular compliance is a multifaceted challenge extending beyond pure logical reasoning.

\section{Appendix: Prompt Constraints Parameters}
\label{sec:appendix}
For each product or service in our evaluation, we systematically defined both tone options and a set of product-specific keywords to guide content generation. The following options were considered:

\begin{itemize}
    \item \textbf{Tone Options:}
    \begin{itemize}
        \item Empathetic and apologetic
        \item Formal and authoritative
        \item Enthusiastic and inspiring
        \item Neutral and objective
    \end{itemize}
    \item \textbf{Product-Specific Keywords:}
    \begin{itemize}
        \item \textbf{Smartphone:}
        \begin{itemize}
            \item performance
            \item innovation
            \item battery life
            \item camera
            \item display
        \end{itemize}
        \item \textbf{Wireless Earbuds:}
        \begin{itemize}
            \item wireless
            \item comfort
            \item noise cancellation
            \item battery
            \item waterproof
        \end{itemize}
        \item \textbf{Savings Account:}
        \begin{itemize}
            \item security
            \item growth
            \item flexibility
            \item FDIC insured
            \item no fees
        \end{itemize}
        \item \textbf{Credit Card:}
        \begin{itemize}
            \item cashback
            \item rewards
            \item no annual fee
            \item convenience
            \item security
        \end{itemize}
        \item \textbf{Meditation App:}
        \begin{itemize}
            \item mindfulness
            \item wellness
            \item relaxation
            \item guided
            \item sleep
        \end{itemize}
        \item \textbf{Fitness Monitor:}
        \begin{itemize}
            \item health
            \item tracking
            \item motivation
            \item progress
            \item wellbeing
        \end{itemize}
        \item \textbf{Meals Subscription:}
        \begin{itemize}
            \item fresh
            \item convenience
            \item variety
            \item nutrition
            \item chef-designed
        \end{itemize}
        \item \textbf{News Streaming:}
        \begin{itemize}
            \item live
            \item global
            \item in-depth
            \item analysis
            \item access
        \end{itemize}
    \end{itemize}
\end{itemize}

During prompt construction, a single tone was randomly selected from the above options, and a subset of keywords for each product was chosen to be either included or avoided, ensuring diversity and control in the generated outputs.

\section{Appendix: Constraints Distributions Charts}
\label{sec:appendix_img}
\begin{figure*}[h]
\includegraphics[width=\linewidth]{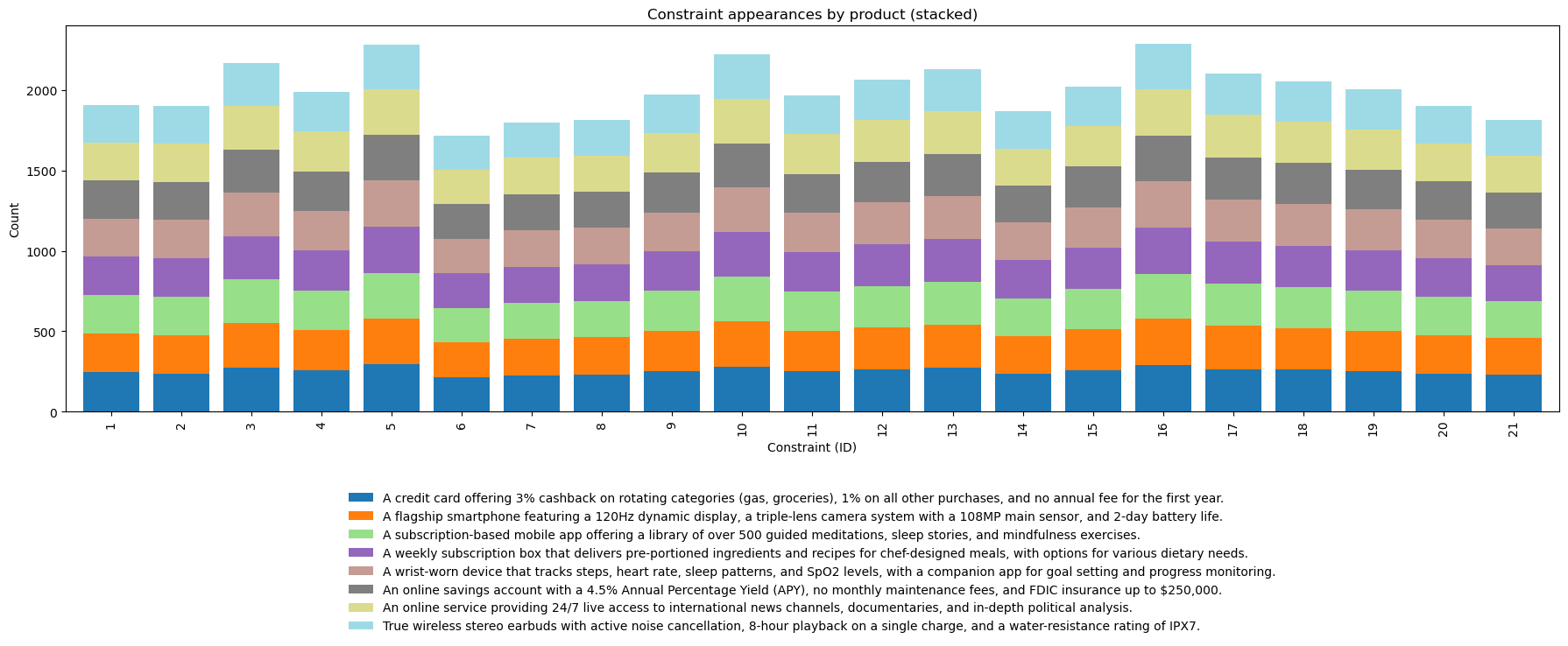}
\caption{Constraints frequency by product/service description.}
\label{fig:product}
\end{figure*}

\begin{figure*}[h]
\includegraphics[width=\linewidth]{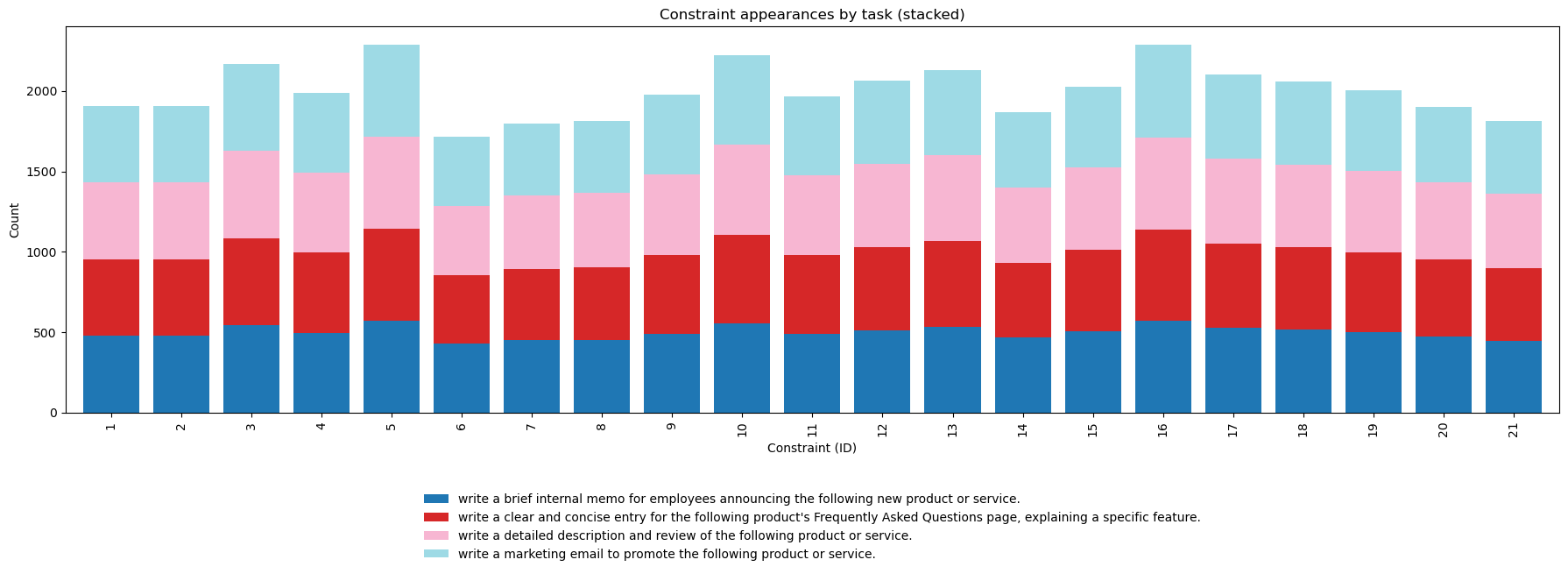}
\caption{Constraints frequency by task.}
\label{fig:task}
\end{figure*}

We report in Figures \ref{fig:product} and \ref{fig:task} additional charts depicting the distribution of constraints by Task and Product/Service description, showing how our dataset is balanced with respect to these aspects. 


\section{Appendix: Evaluation Functions}
\label{sec:appendix_eval_func}

This Section details the rule-based evaluation strategies used to assess a model's compliance with various constraints. For constraints in the Formatting, Lexical, and Syntactic categories, we implemented a suite of automated evaluation functions. These strategies provide a quantitative score, typically between 0.0 (complete failure) and 1.0 (perfect compliance), often calculating partial scores based on the degree of adherence. For the more nuanced Semantic and Business/Legal constraints, evaluation relies on an LLM-as-a-judge approach, as noted in the paper.

\subsection{Formatting Constraints}

\textbf{Paragraph Count.} To evaluate constraints on the number of paragraphs, the text is split by 1+ newline characters. The resulting number of paragraphs is compared against the specified range (e.g., 2-3 paragraphs) to determine a score.

\noindent \textbf{Paragraph Length.} For constraints on paragraph length in sentences, each paragraph is first isolated and then split into individual sentences using regular expressions. The number of sentences is counted to verify it falls within the desired range (e.g., 2-4 sentences).

\noindent \textbf{List Formatting.} To verify list formatting, the evaluation process examines each line of the text to confirm it begins with the specified bullet character, such as a dash (-). The final score is the proportion of lines that correctly follow this format.

\noindent \textbf{Token and Keyword Presence.} For constraints requiring special tokens at specific locations (e.g., <BOC> at the start or <EOC> at the end), the strategy involves a binary check for an exact match at the specified position.

\noindent \textbf{JSON Format.} The text's structure is assessed by attempting to parse it as a JSON object. A perfect score is awarded if parsing is successful. If it fails, a partial score is assigned by analyzing the specific parsing error, granting higher scores for minor syntax issues over fundamental structural flaws.

\subsection{Lexical Constraints}

\noindent \textbf{Readability Level.} To assess adherence to a specific readability score, the Flesch Reading Ease formula is applied to the text. A soft score is then assigned based on how closely the result matches the target range (e.g., 70-80). We rely on the textstat\footnote{\url{https://pypi.org/project/textstat/}} library for the computation of this score.

\noindent \textbf{Keyword Usage.} Compliance is measured by checking for the presence or absence of specified keywords. The text is scanned for a predefined list of terms, and the score is determined by the percentage of required keywords found or the percentage of forbidden keywords that are correctly omitted.

\noindent \textbf{Variable Usage.} The entire text is scanned to confirm the presence of a required variable string (e.g., \{\{FirstName\}\}). A positive score is given if the variable is found.

\subsection{Syntactic Constraints}

\noindent \textbf{Sentence Length Variation.} To assess variability, the text is first segmented into sentences using the NLTK English sentence tokenizer. The word count for each sentence is determined, and the standard deviation of these lengths is calculated. A higher standard deviation, indicating greater variety, results in a higher compliance score.

\noindent \textbf{Maximum Sentence Length.} Each sentence is evaluated individually to ensure it does not exceed a specified word limit. The final score represents the proportion of sentences in the text that successfully meet this length constraint.

\subsection{LLM-as-a-judge}
\label{app:llm_as_judge}

For the more nuanced Semantic and Business/Legal constraints, which resist simple rule-based verification, we employ an LLM-as-a-judge approach. This strategy uses a capable LLM (GPT-4o-mini in our experiments) to score a generated passage's compliance with a specific qualitative constraint. The judge model is prompted to provide a score on a scale of 0 (complete disagreement) to 10 (complete agreement) and is required to output its response in a JSON format, facilitating automated parsing of the score and the reasoning behind it.

The evaluation prompts are dynamically constructed from a set of templates. A general system prompt instructs the model on its role as a strict writing coach. The user prompt contains the specific passage to be evaluated and the rubric corresponding to the constraint being checked. 

\subsubsection{Prompt Templates}
The core components of the prompts are as follows:

\paragraph{System Prompt:}
\begin{verbatim}
You are an expert writing coach acting 
as a fair and strict judge. Your task 
is to evaluate a given passage based 
on a provided rubric.
\end{verbatim}

\paragraph{User Prompt:}
\begin{verbatim}
### Passage ###
{passage}

### Rubric ###
Evaluate the given passage on the
 following criterion on a scale 
 of 0 to 10:
{rubric_name}: {rubric_description} 
(0 = completely disagree, 
2 = somewhat disagree,
5 = neutral, 8 = somewhat agree, 
10 = completely agree).

### Instructions ###
Provide your output only in a JSON 
format with the keys ``reasoning'' and 
``score''.
\end{verbatim}

\subsubsection{Evaluation Rubrics}
The following rubrics were used to evaluate compliance with their respective constraints. Placeholders like \texttt{\{\}} are dynamically populated based on the specific context of the prompt.

\begin{itemize}
    \item \textbf{Positive language:} The given passage uses positive and empowering language (e.g., ``opportunity'', ``benefit'', ``simplify'', etc.) and avoids negative or fear-based terms (e.g., ``problem'', ``risk'', ``failure'', etc.).

    \item \textbf{Specific tone:} The given passage maintains a single, consistent, \texttt{\{tone\}} tone throughout the entire context.

    \item \textbf{Inverted pyramid principle:} The given passage presents the most critical piece of information (e.g., the core answer, main announcement, key takeaway, etc.) in the description \texttt{\{product\_description\}}, before supporting details or secondary context.

    \item \textbf{Internal contradictions:} The given passage does not contain any internal contradictions. All stated facts, arguments, and data points are consistent with each other from start to finish.

    \item \textbf{Supporting evidence:} The given passage avoids making unsupported assertions and provides a reason or piece of evidence within the text for every significant claim, benefit or conclusion.

    \item \textbf{Communication purpose:} The given passage has a clear purpose of communication and directly addresses the primary underlying question of the target audience (e.g., ``How will this help me?'', ``What do I need to know?'', ``Is my problem solved?'', etc.).

    \item \textbf{Precise language:} The given passage uses precision and unambiguous language. It avoids vague terms or phrases that could be misinterpreted by the target audience. All instructions, descriptions or conclusions are explicit and clear in the given passage.

    \item \textbf{Accurate product information:} The given passage only contains accurate product names, features, and numerical data (e.g., prices, percentages, etc.) that can be verified against the product description \texttt{\{product\_description\}}.

    \item \textbf{Substantiated superlatives:} If the passage contains superlatives (e.g., best, greatest, etc.) they are substantiated by verifiable sources. Citations and links are also provided as necessary. If no superlatives are used (i.e. when no claim is made), no substantiation for claims need to be provided.
\end{itemize}

\section{Appendix: Constraint Compliance Metrics}
\label{appendix:ccm}
\begin{table*}[h!]
\small
\centering
\resizebox{\linewidth}{!}{%
\begin{tabular}{p{1.5cm} r r r r r r r}
\toprule
\textbf{Constraint Category} & \textbf{Llama-3.1-8B} & \textbf{Llama-3.3-70B} & \textbf{Qwen3 8B} & \textbf{Mixtral-8x-7B} & \textbf{Deepseek-R1 8B} & \textbf{Gemini 2.5 Flash} &  \textbf{Claude 3.7 Sonnet}  \\
\midrule
Lexical & 0.8704 & 0.8691 & 0.9003 & 0.8698 & 0.8159 & 0.9061 & 0.9064 \\
Semantic & 0.9573 & 0.9588 & 0.9848 & 0.9564 & 0.9550 & 0.9512 & 0.9815 \\
Business/Legal & 0.9095 & 0.8957 & 0.9282 & 0.8822 & 0.8780 & 0.9003 & 0.9105 \\
Syntactic & 0.9967 & 0.9984 & 0.9791 & 0.9802 & 0.9792 & 0.9347 & 0.9942 \\
formatting & 0.9160 & 0.9412 & 0.8168 & 0.8360 & 0.7720 & 0.8419 & 0.8876 \\
\bottomrule
\end{tabular}}
\caption{Single Constraint Compliance (SCC) rates for different constraints categories and models.}
\label{tab:scc_cat_scores}
\end{table*}

In Table \ref{tab:scc_cat_scores}, we report the Single Constraint Compliance (SCC) rates for different models, grouped by constraint category.

\section{Appendix: Pairwise Correlation Metrics}
\label{appendix:pwise}
We report in Tables \ref{tab:llama8b_pwise},  \ref{tab:llama70b_pwise},  \ref{tab:qwen_pwise}, \ref{tab:claude_pwise}, \ref{tab:gemini_pwise}, \ref{tab:mixtral_pwise} and \ref{tab:deepseek_pwise} the detailed correlation analysis results associated with the  $PCC_{c_i, c_j}$ metric.

\begin{table}[H]
\small
\centering
\resizebox{\linewidth}{!}{%
\begin{tabular}{p{3cm} rrr}
\toprule
\textbf{Constraint 1} & \textbf{Constraint 2} & \textbf{Pearson r} & \textbf{p-value} \\
\midrule
Include <BOC> token & Include <ECO> token & 0.322 & 9.98e-33 \\
Include <BOC> token & Structure body into lists & 0.463 & 1.44e-76 \\
Flesch reading ease 70-80 & Substantiate every claim & 0.208 & 2.9e-15 \\
Flesch reading ease 70-80 & Use given keywords & -0.335 & 4.12e-36 \\
Flesch reading ease 70-80 & Keep paragraphs short & 0.353 & 1.94e-44 \\
Flesch reading ease 70-80 & Structure body into lists & 0.239 & 2.14e-20 \\
Avoid given keywords & Substantiate every claim & 0.237 & 8.42e-14 \\
Avoid given keywords & Use given keywords & -0.333 & 2.17e-29 \\
Substantiate every claim & Use given keywords & -0.223 & 9.46e-13 \\
Keep paragraphs short & Keep body short & 0.328 & 1.25e-41 \\
Keep paragraphs short & Vary sentence length & -0.211 & 4.8e-15 \\
Use given tone & Use positive language & 0.329 & 7.74e-32 \\
\bottomrule
\end{tabular}}
\caption{Statistically significant Pearson's correlation coefficients -- correlation coefficient > 0.2 or < -0.2 and p-value < 0.05 -- for pairwise constraint compliance of the Llama 3.1 8B model.}
\label{tab:llama8b_pwise}
\end{table}

\begin{table}[H]
\small
\centering
\resizebox{\linewidth}{!}{%
\begin{tabular}{p{3cm} rrr}
\toprule
\textbf{Constraint 1} & \textbf{Constraint 2} & \textbf{Pearson r} & \textbf{p-value} \\
\midrule
Report Correct Features & Substantiate every claim & 0.205 & 7.15e-13 \\
Report Correct Features & Use inverted pyramid principle & 0.246 & 2.36e-18 \\
Include <BOC> token & Include <ECO> token & 0.225 & 1.95e-16 \\
Flesch reading ease 70-80 & Avoid given keywords & 0.236 & 1.81e-17 \\
Flesch reading ease 70-80 & Use given keywords & -0.278 & 3.37e-25 \\
Flesch reading ease 70-80 & Keep paragraphs short & 0.323 & 3.03e-37 \\
Avoid given keywords & Substantiate every claim & 0.379 & 1.19e-34 \\
Avoid given keywords & Use given keywords & -0.443 & 5.25e-53 \\
Substantiate every claim & Use given keywords & -0.259 & 6.54e-17 \\
Keep paragraphs short & Use Custom Variable & -0.239 & 2e-20 \\
Keep paragraphs short & Keep body short & 0.450 & 1.59e-81 \\
Keep paragraphs short & Vary sentence length & -0.242 & 1.25e-19 \\
\bottomrule
\end{tabular}}
\caption{Statistically significant Pearson's correlation coefficients -- correlation coefficient > 0.2 or < -0.2 and p-value < 0.05 -- for pairwise constraint compliance of the Llama 3.3 70B model.}
\label{tab:llama70b_pwise}
\end{table}

\begin{table}[H]
\small
\centering
\resizebox{\linewidth}{!}{%
\begin{tabular}{p{3cm} rrr}
\toprule
\textbf{Constraint 1} & \textbf{Constraint 2} & \textbf{Pearson r} & \textbf{p-value} \\
\midrule
Report Correct Features & Substantiate every claim & 0.294 & 4.28e-29 \\
Report Correct Features & Avoid contradictions & 0.480 & 1.49e-61 \\
Include <BOC> token & Keep paragraphs short & 0.246 & 2.35e-14 \\
Include <BOC> token & Structure body into lists & 0.480 & 7.21e-61 \\
Include <BOC> token & Respond in JSON & -0.531 & 3.35e-78 \\
Include <BOC> token & Keep body short & 0.514 & 1.06e-76 \\
Flesch reading ease 70-80 & Avoid given keywords & 0.357 & 1.25e-44 \\
Flesch reading ease 70-80 & Use given keywords & -0.337 & 1.06e-39 \\
Flesch reading ease 70-80 & Keep body short & 0.209 & 2.4e-15 \\
Avoid given keywords & Substantiate every claim & 0.286 & 3.4e-24 \\
Substantiate every claim & Use inverted pyramid principle & 0.226 & 1.46e-19 \\
Use given keywords & Keep paragraphs short & -0.227 & 3.21e-15 \\
Use given keywords & Keep body short & -0.214 & 5.47e-15 \\
Keep paragraphs short & Structure body into lists & 0.453 & 1.4e-51 \\
Keep paragraphs short & Keep body short & 0.783 & 7.1e-223 \\
Use given tone & Clear purpose & 0.281 & 1.87e-24 \\
Use given tone & Avoid contradictions & 0.490 & 9.52e-57 \\
Structure body into lists & Respond in JSON & -0.250 & 1.32e-17 \\
Structure body into lists & Keep body short & 0.585 & 2.42e-113 \\
Use Custom Variable & Vary sentence length & 0.205 & 1.12e-12 \\
Respond in JSON & Keep body short & -0.264 & 8.08e-21 \\
Use inverted pyramid principle & Avoid contradictions & 0.300 & 1.06e-27 \\
Clear purpose & Avoid contradictions & 0.324 & 1.21e-27 \\
\bottomrule
\end{tabular}}
\caption{Statistically significant Pearson's correlation coefficients -- correlation coefficient > 0.2 or < -0.2 and p-value < 0.05 -- for pairwise constraint compliance of the Qwen 3 8B model.}
\label{tab:qwen_pwise}
\end{table}

\begin{table}[H]
\small
\centering
\resizebox{\linewidth}{!}{%
\begin{tabular}{p{3cm} rrr}
\toprule
\textbf{Constraint 1} & \textbf{Constraint 2} & \textbf{Pearson r} & \textbf{p-value} \\
\midrule
Report Correct Features & Substantiate every claim & 0.443 & 6.27e-59 \\
Report Correct Features & Use unambiguous language & 0.515 & 3.43e-82 \\
Include <BOC> token & Include <ECO> token & 0.542 & 1.61e-100 \\
Include <BOC> token & Avoid given keywords & 0.238 & 1.57e-14 \\
Include <BOC> token & Structure body into lists & 0.451 & 1.95e-72 \\
Include <BOC> token & Respond in JSON & 0.625 & 2.15e-110 \\
Include <ECO> token & Flesch reading ease 70-80 & 0.205 & 5.93e-15 \\
Include <ECO> token & Structure body into lists & 0.338 & 6.99e-39 \\
Include <ECO> token & Respond in JSON & 0.295 & 7.15e-27 \\
Include <ECO> token & Use inverted pyramid principle & 0.236 & 8.31e-18 \\
Flesch reading ease 70-80 & Structure body into lists & 0.416 & 3.82e-62 \\
Flesch reading ease 70-80 & Vary sentence length & -0.202 & 2.32e-14 \\
Substantiate every claim & Use inverted pyramid principle & 0.232 & 2.78e-15 \\
Substantiate every claim & Clear purpose & 0.213 & 3.81e-14 \\
Keep paragraphs short & Keep body short & 0.404 & 1.54e-64 \\
Keep paragraphs short & Vary sentence length & -0.218 & 4.63e-16 \\
Use given tone & Use inverted pyramid principle & 0.219 & 4.34e-14 \\
Structure body into lists & Respond in JSON & 0.222 & 2.72e-15 \\
Use inverted pyramid principle & Use unambiguous language & 0.385 & 1.2e-42 \\
Keep body short & Vary sentence length & -0.237 & 6.43e-19 \\
Clear purpose & Avoid contradictions & 0.226 & 2.34e-18 \\
Clear purpose & Use positive language & 0.676 & 2e-183 \\
Avoid contradictions & Use unambiguous language & -0.385 & 4.84e-46 \\
Use positive language & Use unambiguous language & 0.291 & 6.57e-25 \\
\bottomrule
\end{tabular}}
\caption{Statistically significant Pearson's correlation coefficients -- correlation coefficient > 0.2 or < -0.2 and p-value < 0.05 -- for pairwise constraint compliance of the Claude 3.7 Sonnet model.}
\label{tab:claude_pwise}
\end{table}

\begin{table}[H]
\small
\centering
\resizebox{\linewidth}{!}{%
\begin{tabular}{p{3cm}rrr}
\toprule
\textbf{Constraint 1} & \textbf{Constraint 2} & \textbf{Pearson r} & \textbf{p-value} \\
\midrule
Report Correct Features & Use inverted pyramid principle & 0.295 & 4.72e-26 \\
Report Correct Features & Clear purpose & 0.812 & 2.96e-304 \\
Report Correct Features & Avoid contradictions & 0.856 & 6.72e-315 \\
Report Correct Features & Use positive language & 0.672 & 1.64e-154 \\
Include <BOC> token & Include <ECO> token & 0.346 & 1.03e-37 \\
Include <BOC> token & Keep paragraphs short & 0.562 & 3.69e-111 \\
Include <BOC> token & Structure body into lists & 0.737 & 1.41e-243 \\
Include <BOC> token & Keep body short & 0.670 & 9.31e-169 \\
Include <ECO> token & Structure body into lists & 0.303 & 5.72e-31 \\
Include <ECO> token & Keep body short & 0.206 & 2.4e-16 \\
Flesch reading ease 70-80 & Structure body into lists & 0.233 & 2.44e-19 \\
Flesch reading ease 70-80 & Keep body short & 0.224 & 9.42e-21 \\
Substantiate every claim & Use inverted pyramid principle & 0.224 & 3.47e-14 \\
Substantiate every claim & Use unambiguous language & 0.257 & 1.62e-18 \\
Keep paragraphs short & Structure body into lists & 0.583 & 2.23e-127 \\
Keep paragraphs short & Keep body short & 0.705 & 7.01e-242 \\
Keep paragraphs short & Vary sentence length & -0.204 & 3.17e-14 \\
Use given tone & Clear purpose & -0.214 & 6.3e-15 \\
Use given tone & Avoid contradictions & 0.234 & 1.78e-17 \\
Use given tone & Use positive language & 0.292 & 5.01e-25 \\
Structure body into lists & Keep body short & 0.796 & 0 \\
Structure body into lists & Vary sentence length & -0.215 & 1.15e-14 \\
Keep body short & Vary sentence length & -0.225 & 4.83e-17 \\
Clear purpose & Avoid contradictions & 0.640 & 1.57e-168 \\
Clear purpose & Use positive language & 0.715 & 3.33e-213 \\
Avoid contradictions & Use positive language & 0.915 & 0 \\
\bottomrule
\end{tabular}}
\caption{Statistically significant Pearson's correlation coefficients -- correlation coefficient > 0.2 or < -0.2 and p-value < 0.05 -- for pairwise constraint compliance of the Gemini Flash 2.5 model.}
\label{tab:gemini_pwise}
\end{table}

\begin{table}[H]
\small
\centering
\resizebox{\linewidth}{!}{%
\begin{tabular}{p{3cm} rrr}
\toprule
\textbf{Constraint 1} & \textbf{Constraint 2} & \textbf{Pearson r} & \textbf{p-value} \\
\midrule
Report Correct Features & Include <ECO> token & 0.218 & 8.07e-15 \\
Report Correct Features & Substantiate every claim & 0.404 & 3.51e-48 \\
Report Correct Features & Use given tone & 0.246 & 2.86e-17 \\
Report Correct Features & Use inverted pyramid principle & 0.324 & 1.77e-31 \\
Report Correct Features & Clear purpose & 0.344 & 1.62e-37 \\
Report Correct Features & Avoid contradictions & 0.203 & 1.12e-11 \\
Report Correct Features & Use positive language & 0.267 & 1.12e-20 \\
Report Correct Features & Use unambiguous language & 0.343 & 1.69e-34 \\
Report Correct Features & Vary sentence length & 0.222 & 5.84e-14 \\
Include <ECO> token & Flesch reading ease 70-80 & -0.321 & 1.2e-35 \\
Include <ECO> token & Substantiate every claim & 0.280 & 6.39e-22 \\
Include <ECO> token & Use given keywords & 0.294 & 1.43e-25 \\
Include <ECO> token & Use given tone & 0.371 & 1.2e-45 \\
Include <ECO> token & Structure body into lists & 0.246 & 8.69e-21 \\
Include <ECO> token & Respond in JSON & 0.326 & 9.77e-33 \\
Include <ECO> token & Use inverted pyramid principle & 0.377 & 7.2e-45 \\
Include <ECO> token & Clear purpose & 0.458 & 2.5e-80 \\
Include <ECO> token & Avoid contradictions & 0.226 & 6.08e-18 \\
Include <ECO> token & Use positive language & 0.286 & 1.23e-26 \\
Include <ECO> token & Use unambiguous language & 0.394 & 9.98e-50 \\
Include <ECO> token & Vary sentence length & 0.353 & 6.23e-35 \\
Flesch reading ease 70-80 & Avoid given keywords & 0.277 & 9.41e-24 \\
Flesch reading ease 70-80 & Use given keywords & -0.320 & 3.19e-33 \\
Flesch reading ease 70-80 & Use given tone & -0.221 & 1.59e-17 \\
Flesch reading ease 70-80 & Clear purpose & -0.264 & 2.38e-23 \\
Flesch reading ease 70-80 & Use unambiguous language & -0.203 & 7.99e-14 \\
Avoid given keywords & Substantiate every claim & 0.208 & 5.64e-11 \\
Avoid given keywords & Use given keywords & -0.320 & 4.31e-27 \\
Substantiate every claim & Use given tone & 0.242 & 7.5e-18 \\
Substantiate every claim & Use inverted pyramid principle & 0.306 & 6.27e-26 \\
Substantiate every claim & Clear purpose & 0.399 & 1.34e-48 \\
Substantiate every claim & Use positive language & 0.233 & 2.42e-14 \\
Substantiate every claim & Use unambiguous language & 0.328 & 6.65e-30 \\
Substantiate every claim & Vary sentence length & 0.232 & 6.36e-15 \\
Use given keywords & Use given tone & 0.283 & 1.89e-22 \\
Use given keywords & Clear purpose & 0.312 & 2.43e-30 \\
Use given keywords & Avoid contradictions & 0.217 & 1.04e-14 \\
Use given keywords & Use positive language & 0.294 & 1.26e-23 \\
Use given keywords & Use unambiguous language & 0.303 & 7.81e-27 \\
Use given keywords & Vary sentence length & 0.273 & 2.66e-19 \\
Use given tone & Structure body into lists & 0.200 & 3.02e-13 \\
Use given tone & Use inverted pyramid principle & 0.439 & 3.61e-56 \\
Use given tone & Clear purpose & 0.525 & 2.12e-93 \\
Use given tone & Avoid contradictions & 0.548 & 1.05e-102 \\
Use given tone & Use positive language & 0.640 & 3.06e-140 \\
Use given tone & Use unambiguous language & 0.565 & 1.39e-102 \\
Use given tone & Vary sentence length & 0.463 & 1.64e-63 \\
Structure body into lists & Clear purpose & 0.238 & 1.72e-19 \\
Respond in JSON & Clear purpose & 0.204 & 1.05e-13 \\
Use inverted pyramid principle & Clear purpose & 0.514 & 7.18e-97 \\
Use inverted pyramid principle & Avoid contradictions & 0.420 & 2.5e-52 \\
Use inverted pyramid principle & Use positive language & 0.487 & 3.87e-76 \\
Use inverted pyramid principle & Use unambiguous language & 0.477 & 2.52e-67 \\
Use inverted pyramid principle & Vary sentence length & 0.385 & 1.16e-43 \\
Clear purpose & Avoid contradictions & 0.377 & 1.01e-50 \\
Clear purpose & Use positive language & 0.567 & 5.01e-117 \\
Clear purpose & Use unambiguous language & 0.575 & 5.29e-121 \\
Clear purpose & Vary sentence length & 0.483 & 1.88e-73 \\
Avoid contradictions & Use positive language & 0.482 & 7.98e-71 \\
Avoid contradictions & Use unambiguous language & 0.411 & 7.26e-53 \\
Avoid contradictions & Vary sentence length & 0.241 & 1.5e-16 \\
Use positive language & Use unambiguous language & 0.497 & 3.98e-76 \\
Use positive language & Vary sentence length & 0.400 & 1.16e-47 \\
Use unambiguous language & Vary sentence length & 0.329 & 2.5e-29 \\
\bottomrule
\end{tabular}}
\caption{Statistically significant Pearson's correlation coefficients -- correlation coefficient > 0.2 or < -0.2 and p-value < 0.05 -- for pairwise constraint compliance of the Mixtral 8x 7B model.}
\label{tab:mixtral_pwise}
\end{table}

\begin{table}[H]
\small
\centering
\resizebox{\linewidth}{!}{%
\begin{tabular}{p{3cm} rrr}
\toprule
\textbf{Constraint 1} & \textbf{Constraint 2} & \textbf{Pearson r} & \textbf{p-value} \\
\midrule
Report Correct Features & Substantiate every claim & 0.341 & 3.3e-39 \\
Report Correct Features & Use inverted pyramid principle & 0.516 & 7.21e-108 \\
Report Correct Features & Avoid contradictions & 0.203 & 3.38e-11 \\
Include <BOC> token & Include <ECO> token & 0.222 & 2.8e-15 \\
Avoid unsubstantiated superlatives & Substantiate every claim & 0.299 & 1.78e-26 \\
Flesch reading ease 70-80 & Use given keywords & -0.303 & 6.27e-32 \\
Flesch reading ease 70-80 & Keep body short & 0.260 & 4.21e-23 \\
Substantiate every claim & Use inverted pyramid principle & 0.280 & 1.59e-29 \\
Use given keywords & Use Custom Variable & 0.219 & 5.24e-14 \\
Use given tone & Clear purpose & -0.253 & 6.12e-20 \\
Use given tone & Use positive language & -0.231 & 2.29e-15 \\
Use given tone & Use unambiguous language & -0.229 & 2e-18 \\
Respond in JSON & Keep body short & -0.272 & 4.06e-22 \\
Clear purpose & Use positive language & 0.792 & 2.99e-245 \\
Clear purpose & Use unambiguous language & 0.866 & 0 \\
Avoid contradictions & Use unambiguous language & 0.288 & 5.03e-23 \\
Use positive language & Use unambiguous language & 0.691 & 1.01e-186 \\
\bottomrule
\end{tabular}}
\caption{Statistically significant Pearson's correlation coefficients -- correlation coefficient > 0.2 or < -0.2 and p-value < 0.05 -- for pairwise constraint compliance of the Deepseek R1 8B model.}
\label{tab:deepseek_pwise}
\end{table}

\end{document}